\documentclass[runningheads]{llncs}

\authorrunning{H. N. Adorna}
\titlerunning{Configuration graph and SN P Systems}

\usepackage{tikz}
\usetikzlibrary{arrows,shapes,automata,petri}

\usetikzlibrary{shapes.geometric, arrows}

\tikzstyle{arrow} = [thick,->,>=stealth]
\tikzstyle{neuron} = [rectangle, rounded corners, minimum width=2cm, minimum height=1cm, text centered, text width=2cm, draw=black]
\tikzstyle{o} = [rectangle, minimum width=0.15cm, minimum height=0.15cm, draw=white,fill=white]

\usepackage[intlimits]{amsmath}
\usepackage{amsmath}
\usepackage{amssymb}
\usepackage{mathtools}
\usepackage{graphicx}
\usepackage[colorinlistoftodos]{todonotes}
\usepackage{enumerate}
\usepackage{verbatim}
\usepackage{epsf}
\usepackage{latexsym}
\usepackage{cite}
\usepackage{setspace}
\usepackage{hyperref,multirow}
\usepackage{zed-csp}
\usepackage{pdflscape}

\usepackage{color}
\usepackage{todonotes}
\usepackage{amssymb,amsmath}
\usepackage{multirow}
\usepackage{pdflscape}
\usepackage{afterpage}

\usepackage{cite}
\usepackage{authblk}
\usepackage[T1]{fontenc}
\usepackage[utf8]{inputenc}

\newtheorem{observation}{Observation}

\newcommand{\ra}{\rightarrow}

\newcommand{\myqed}{\begin{flushright}$\blacksquare$\end{flushright}}

\title{Properties of SN P system and its Configuration Graph}
\author{
Henry N. Adorna
}
\institute{
Department of Computer Science (Algorithm and Complexity)\\
University of the Philippines Diliman\\
         1101 Quezon City, Philippines\\
E-mail: {{\tt hnadorna@up.edu.ph
}}
}

\begin{document}

\maketitle

\begin{abstract}
Several studies have been reported in the literature about SN P system and its variants.  Often, the results provide universality of various variants and the classes of languages that these variants generate and recognize.  The state of SN P system is its configuration.  We refer to our previous result on reachability of configuration as the {\it Fundamental state equation for SN P system.} This paper provides a preliminary investigation on the behavioral and structural properties of SN P system without delay that depend primarily to this fundamental state equation.    Also, we introduce the idea of configuration graph $CG_{\Pi}$ of an SN P system $\Pi$  without delay to characterize behavioral properties of $\Pi$ with respect to  $CG_{\Pi}.$   The matrix $M_{\Pi}$ of an SN P system $\Pi$ without delay is used to characterize structural properties of $\Pi.$  
\end{abstract}
{\bf Keywords: } Membrane Computing, Spiking Neural P System, Matrices, Configuration Graphs, Behavioral and Structural Properties

%----------------------------------------
\section{Introduction}
%----------------------------------------
In the 2006, Spiking Neural P system \cite{ionescu2006} was introduced as a neural-like P system that is an additional to cell-like and tissue-like types of P system models in  Membrane computing \cite{paun2000}.  Several results have been reported about SN P system including its different variants, particularly proving universality and characterization of languages they recognize and generate \cite{chen2008, ibarra2010}.  Others considered  application areas \cite{zhang2017, frisco2014, ciobanu2006} and simulations of SN P systems {\it in silico} \cite{carandang2017, delacruz2018b, jimenez2018, delamor2017}. Survey of results on SN P systems (until 2016) is reported in \cite{pan2016}.

In the 2010, Zeng et al \cite{zeng-etal2010} introduced a matrix representation of a Spiking Neural P (SN P) system (without delay).  Several {\it in silico} implementations of SN P systems \cite{carandang2017, delacruz2018b, jimenez2018, delamor2017} are based on the matrix representation of  \cite{zeng-etal2010} including the matrix representation of some variants of P systems \cite{delacruz2018, juayong2011}.   

Recently, \cite{adorna2019}, defined reachability of configuration of SN P systems  using the solvability of the matrix equation defined in \cite{zeng-etal2010}.  This  equation plays significant and fundamental role in this paper, so that we refer to it as follows,

\begin{quote}
\noindent{\bf Fundamental state equation for SN P systems}

Let $\Pi$ be an SN P system.  Let $k \in \mathbb{Z} \cup  \{0\}.$  Then
\begin{equation}\label{eq1}
C^{(k+1)} = C^{(k)} + Sp^{(k)} \cdot M_{\Pi}
\end{equation}
where $C^{(k+1)}$ and  $C^{(k)}$ are the $(k+1)^{th}$ and $k^{th}$ configurations, respectively.  $Sp^{(k)}$ is a valid spiking vector obtained from $C^{(k)},$ and $M_{\Pi}$ is the matrix representation of $\Pi.$ 
\end{quote}

%In the 2011, Ibo et al \cite{Ibo2011} reported on the periodicity as a dynamical aspect of generative SN P system $\Pi.$  The paper \cite{Ibo2011} considered a square matrix representation of $M_{\Pi}.$  It was shown that if $\Pi$ is periodic, then the determinant of $M_{\Pi}$ is zero, where being periodic means existence of configurations $C^{(k)}$ and $C^{(p)},$ such that $C^{(k)} = C^{(p)},$ where $k \not = p.$

 Also, in \cite{adorna2019},  $\text{\bf Struct-}M_{\Pi}$ is defined as a square matrix associated with $\Pi$ that considers only (directed) connectivity of neurons by synapses.  % and reduces all the rules in all neurons to something like $a \rightarrow a.$  This leaves all diagonal entries to be $-1,$ while an entry $1$ means the neurons are linked by a directed edge (or synapse), else $0.$  %Notice, that this is another way fo putting the results in \cite{Ibo2011}. 
$\Pi$ has a cycle, if $\text{row-rank}(\text{\bf Struct-}M_{\Pi}) < m,$ where $m$ is the number of neurons in $\Pi.$ 
This idea of considering only the graph structure of SN P systems was mentioned earlier in \cite{Ibo2011} in considering periodicity in SN P system. 

 In this paper, we look closely into the graph representation of computation by SN P systems.  We investigate on the behavioral and structural properties of SN P systems using this graph representation.  We introduce and define the concept of a configuration graph $CG_{\Pi}$ for SN P system $\Pi$ without delay based on equation (\ref{eq1}).   This concept is similar to the derivation tree that describes the computation of a formal grammar and that of the computation trees associated with P system described in \cite{franco2005}.  The configuration graph $CG_{\Pi}$ of SN P system $\Pi$ without delay is a tuple $CG_{\Pi} = (V,E)$ where $V(CG_{\Pi})$ is the set of all reachable configurations of $\Pi$ and $E(CG_{\Pi})$ is the set of pairs $(C^{(i)}, C^{(j)})$  such that $C^{(j)} = C^{(i)} + Sp^{(i)} \cdot M_{\Pi},$ for some valid spiking vector $Sp^{(i)}.$   The graph $CG_{\Pi}$ is an edge-labeled directed graph whose labels are the valid spiking vectors. We do initial investigation on some petri net-like properties of SN P system $\Pi$ as defined in \cite{cabarle2013, marian2020, murata1989, reisig1998, david-etal2010} with respect to its configuration graph $CG_{\Pi}.$ These properties of SN P system $\Pi$ that depends on its (fixed) initial configuration will be distinguished as {\it behavioral properties} as opposed to those that only rely on the topology or connectivity of neurons by synapses.  The properties of SN P systems that do not depend on (a fixed) initial configurations are called {\it structural properties.}  Structural properties of SN P system $\Pi$ are investigated here using its matrix representation $M_{\Pi}.$
 
 This paper is a preliminary work on the studies of behavioral and structural properties of SN P systems without delay.  This continues the explorations of the matrix representation of SN P systems started in \cite{zeng-etal2010, adorna2019}.  We provided required definitions and concepts needed for this work in Section \ref{prelim}. The idea of configuration graph is presented in Section \ref{configgraph}.  Petri net-like properties of SN P systems without delay are defined in Section \ref{properties}.  Then we proceed to Section \ref{snp+configgraph} to present our results characterizing behavioral and structural properties of SN P system $\Pi$ without delay via configuration graphs and matrix representation of $\Pi,$ respectively.  We finish the paper with some remarks and observations in Section \ref{finalremarks}.

%---------------------------------------
\section{Preliminaries} \label{prelim}
%---------------------------------------
%-definition and descriptions of SN P systems without delay
We assume the readers to be familiar with membrane computing and its neural-like type of P system as presented in \cite{paun2002}, and \cite{paun2010}, respectively.   The basic understanding of linear algebra, particularly, on vectors and matrix  operations will be useful.
 
\begin{definition}\label{def:SNP}
	An \emph{SN P system without delay} of degree $m$, $m\ge 1$, is a tuple 
$$\Pi=(O,\sigma_1,\dots, \sigma_m, syn, in, out),$$ where
\begin{itemize}
\item $O=\{a\}$ is a singleton alphabet ($a$ is called spike);
\item $\sigma_i, 1\le i \le m,$ are neurons, $\sigma_i=(n_i,R_i), 1\le i \le m,$ where
\begin{itemize}
\item $n_i\ge 0$ is the number of spikes in $\sigma_i$;
\item $R_i$ is a finite set of rules of the following forms:
\begin{itemize}
\item (\emph{Type (1); spiking rules})\\
$E/a^c \ra a^p;$ where $E$ is a regular expression over $\{a\}$, and $c\ge 1,$ $p\ge 1,$  such that $c\ge p;$
\item (\emph{Type (2); spiking rules})\\
$a^s\ra \lambda,$ for $s\ge1,$ such that for each rule $E/a^c \ra a^p$ of type (1) from $R_i$, $a^s\notin L(E);$ 
\end{itemize}
\end{itemize}
\item $syn = \{(i,j)| 1\le i, j \le m, i\ne j\}$ (synapses between neurons);
\item $in, out \in \{1,\dots. m\}$ indicate the input and output neurons respectively.
\end{itemize}
\end{definition}

The SN P system $\Pi$ computes by applying one rule at a time from each neuron, but the executions happen simultaneously at the same time.  In this work, all SN P systems are only those without delay, unless stated otherwise.

\begin{definition}{\bf (Configuration Vector of SN P systems $\Pi$ without delay)}\\
A {\bf configuration} of SN P system $\Pi$ is an $m$-size vector of integers  $$C = (a_1, a_2, \ldots , a_m),$$ where $a_j$, $1\le j\le m,$ represents the number of spikes in neuron $\sigma_j.$
  
A {\bf configuration at time $k,$ $k \geq 0,$} of an SN P system $\Pi,$ as above, is a vector $$C^{(k)} = (a_1^{(k)}, a_2^{(k)}, \ldots , a_m^{(k)}),$$ where $a_j^{(k)} \in \mathbb{Z}^{+} \cup \{0\},$  $1\le j\le m$, is the number of spikes present at time $k$ in neuron $\sigma_j.$  

The vector $C^{(0)}=(a_1^{(0)}, a_2^{(0)} , \ldots , a_m^{(0)})$ is the initial configuration vector of SN P system $\Pi,$ where $a_j^{(0)},$ $1\le j\le m,$ represents the initial number of spikes in neuron $\sigma_j.$ 
\end{definition}

We say a rule $r_x\in R_j,$ $1\le j\le m,$ of neuron $\sigma_j$ is {\bf applicable} at time $k,$ $k\ge 0,$ if and only if the amount of spikes in the neuron, $a_j^{(k)},$ satisfies $E_x$  (or  $a_j^{(k)} \in L(E_{x})).$   At some time $k,$ if we have in neuron $\sigma_j,$ $a_j^{(k)} \in L(E_x) \cap L(E_y),$ for some rules $r_x,$ and $r_y,$  $x \not = y,$ then one of the two rules will be chosen to be applied. This is how the nondeterminism of the system is realized.  Whereas several neurons with their chosen applicable rules can fire (or spike) simultaneouly at time $k,$ demonstrating parallelism. 

A rule of type 2 from $R_j$ removes spike(s) from the neuron at some time $k$ when applied.  Such a rule could only be applied if and only if the number of spikes in $\sigma_j$ is exactly the amount of spikes it needs to be applied.  In particular, each rule $E/a^c \rightarrow a^p ; 0$ of type (1) from $R_i,$ if $a^s \rightarrow \lambda \in R_i,$ then  $a^s \not \in L(E).$
Formally, if there exists a rule $a^s \rightarrow  \lambda \in R_j,$ then there cannot exist any rule $E / a^c \rightarrow a^p$ of type $1$ in $R_j$ such that $a^s \in L(E).$

In specifying rules in SN P systems, we follow the standard convention of simply not specifying $E$ whenever the left-hand side of the rule is equal to $E.$ 

The sequence of configurations defines a {\bf computation} of the system. We say that a computation {\bf halts} or reaches a  {\bf halting configuration} if it reaches a configuration where no more applicable rules are available.  We could represent the output of the system either as the {\bf number of steps lapse between the first two spikes} of the designated output neuron or as {\bf spike train.}  The sequence of spikes made by the system until the system reaches a halting configuration is called  {\bf spike train.}

For an SN P system, $\Pi,$ with $m$ neurons, as introduced in Definition \ref{def:SNP}, we define below  a matrix $M_{\Pi}.$   In the sequel the rules of such an SN P system will be denoted $r_i,$ $1\leq i\leq n$, where $n$ is the total number of rules.

%-matrix representation of SN P systems without delay

\begin{definition}\label{matrix-SNP-d} {\bf (Spiking transition matrix)} \\
Let $\Pi$ be an SN P system with the total number of rules $n$  and $m$ neurons. The {\bf spiking transition matrix} of $\Pi$ is $M_{\Pi} = [ b_{ij} ]_{n \times m},$
where

\[ b_{ij}  =
  \begin{cases}
    -c_i,       & \quad \text{if the left-hand side rule of } r_i  \text{ in } \sigma_j \text{ is } a^{c_i}\\
    & \quad \text{ (} c_i \text{ spikes are consumed)}\\
   ~ ~ ~ p_i,  & \quad \text{if the right-hand side of the rule } r_i \text{ in } \sigma_s \\
   & \quad (s \neq j \text{ and } (s,j) \in syn) \text{ is } a^{p_i}\\ 
 ~ ~ ~ 0, & \quad \text{otherwise}  \end{cases}
\]

\end{definition}

The matrix $M_{\Pi}$ is (almost) a natural representation of the SN P system $\Pi.$  Each row $i$, $1\le i\le n$, corresponds to a rule $r_i: E_i / a^{c_i} \rightarrow a^{p_i}$ in some neuron $\sigma_j,$ with $b_{ij},$ $1\le j\le m,$ defined as above. Each column $j$, $1\le j\le m,$ corresponds to a neuron $\sigma_j$ and (i) for each rule belonging to $\sigma_j$, $r_i: E_i / a^{c_i} \rightarrow a^{p_i},$ for some $i,$ $1\le i\le n,$ $b_{ij}=-c_i$; (ii) for each rule $r_i: E_i / a^{c_i} \rightarrow a^{p_i}$ belonging to $\sigma_s$ for some $s,$ $1\le s\le m,$ such that $s \neq j$ and $(s,j)\in syn$, $b_{ij}=p_i;$ (iii) otherwise, $b_{ij}$ equal to 0.

\begin{example}\label{example1}  {\bf An SN P system for $\mathbb{N} - \{1\}$} \cite{zeng-etal2010}.\\
Let $\Pi = (\{a\}, \sigma_1, \sigma_2, \sigma_3, syn , out),$ where $\sigma_1 = ( 2, R_1),$ with $R_1 = \{ a^2 / a \rightarrow a,  a^2 \rightarrow a\};$  $\sigma_2 = (1, R_2),$ with $R_2 = \{ a \rightarrow a\};$ and $\sigma_3 =(1, R_3),$ with $R_3 = \{ a \rightarrow a, a^2 \rightarrow \lambda \};$ $syn = \{(1,2), (1,3), (2,1), (2,3)\};$  $out = \sigma_3.$
\end{example}

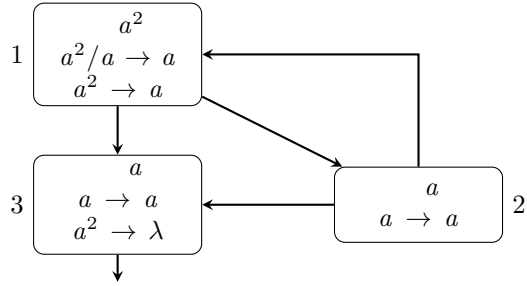
\begin{figure}[h] \label{snp1}
\begin{center}
\begin{tikzpicture}[node distance=2cm]

\node (sigma1) [neuron][label=left:$1$] {~~ $a^2$ ~~~  $a^2 / a \rightarrow a$ $a^2 \rightarrow a$ };
\node (sigma3) [neuron, below of=sigma1, label=left:$3$]{~~~ $a$ ~~~ $a \rightarrow a$ $a^2 \rightarrow \lambda$};
\node (out) [o, below of=sigma3, yshift=0.85cm]{};

\node (sigma2) [neuron, right of=sigma3, xshift=2cm, label=right:$2$]{~~ $a$ ~~~ $a \rightarrow a$};

\node (la) [o, above of=sigma1, xshift=-2cm, yshift=-1cm]{$\Pi \colon$};

\draw [arrow] (sigma1) -- (sigma2);
\draw [arrow] (sigma2) |-  (sigma1);
\draw [arrow] (sigma1) -- (sigma3);
\draw [arrow] (sigma2) -- (sigma3);
\draw [arrow] (sigma3) -- (out);
\end{tikzpicture}
\end{center}
\caption{SN P system $\Pi$ without delay that generates the set $\mathbb{N} -\{1\}.$}
\end{figure}

The SN P system in Example \ref{example1}, generating $\mathbb{N} - \{1\},$ has $n=5$ rules and $m=3$ neurons and its spiking transition matrix is
$
M_{\Pi} = \left (
 \begin{matrix} 
  -1& 1 & 1 \\
  -2 & 1 & 1 \\
  1 & -1 & 1 \\ 
  0 & 0 & -1 \\
  0 & 0 & -2  
 \end{matrix}
\right ).
$  

Some more details regarding the properties of this matrix, $M_{\Pi},$ are presented in \cite{zeng-etal2010, adorna2019}. 
%%\begin{remark}\label{STM}
%%In the spiking transition matrix one can identify all the elements of an SN P system, apart from the initial spikes %and the regular expressions associated with rules.
%%\end{remark}

The initial configuration of $\Pi,$ introduced in Example \ref{example1},  is $C^{0)} = (2,1,1).$

\begin{definition}\label{SpikingVector}{\bf (Valid spiking vector)}\\
Let $C^{(k)} = (a_1^{(k)}, a_2^{(k)}, \ldots , a_m^{(k)}),$ be the current configuration of an SN P system, $\Pi,$ with  a total of $n$ rules and $m$ neurons, at time $k.$  Assume a total order
%% $d:$ $1,2, \ldots , n$ 
is given for all $n$ rules of $\Pi, $ so the rules can be referred to as $r_1, r_2, \ldots, r_n.$  We denote a {\bf valid spiking vector} by
$$Sp^{(k)} = (sp_1^{(k)}, sp_2^{(k)}, \ldots  , sp_n^{(k)}),$$ 
where
\begin{enumerate}  
\item for each neuron $\sigma_j,$ $1\le j \le m,$ if from all possible applicable rules of $R_j$  to $a_j^{(k)}$ at time $k,$ a unique $r_i \in R_j$ is selected to be used, then  $sp^{(k)}_i = 1;$ 
\item for all the rules $r_i$, $1\le i\le n,$ $r_i\in R_1\cup\dots\cup R_m,$ that are not applicable at time $k$ or are not selected, if applicable, then $sp^{(k)}_i = 0.$ 
\end{enumerate}
\end{definition}

One can observe that $\sum_{i=1}^{n} sp^{(k)}_i \leq m.$ 

We denote by $Sp^{(0)} = (sp_1^{(0)}, sp_2^{(0)}, \ldots  , sp_n^{(0)}),$ the {\bf initial valid spiking vector} with respect to $C^{(0)}$ of $\Pi.$   Note that $Sp^{(0)}$ need not be unique (see Example \ref{example1}).

Definition \ref{SpikingVector} implies that $Sp^{(k)} = (sp_1^{(k)}, sp_2^{(k)}, \ldots  , sp_n^{(k)}) %\in \mathbb{Z}_{2}^{n},
$ is a vector such that %is, 
$sp^{(k)}_i \in 
%%\mathbb{Z}_{2},
\{0,1\},$ $1\le i\le n.$    The vector $Sp^{(k)}$ indicates which rules must be fired (or used) at time $k$ given the current configuration $C^{(k)}$ of $\Pi.$  
%%Moreover, if a neuron contains at least two rules and both rules could be applied at some time $k,$ then only one of the two rules must be selected to fire or spike at time $k.$   This is where non-determinism is demonstrated in the system.  
A vector is not a valid spiking vector if it does not satisfy Definition \ref{SpikingVector}.
In our example, the initial spiking vector with respect to $C^{(0)}$ could be either $Sp^{(0)} = (1,0,1,1,0)$ or $(0,1,1,1,0).$   The vector $(1,1,1,1,0)$ is not a valid spiking vector.

%%Applying the %allowable 
%%rules as indicated by $Sp^{(k)}$ at time $k,$ would change the configuration from $C^{(k)}$ to $C^{(k+1)}.$  This means that neurons of  $\Pi$ in the next configuration would either gain or loss some spikes at time $k+1.$  In particular, we could represent such so-called {\bf transition net gain or loss} of $\Pi$ at time $k$ as vector $NG^{(k)}=C^{(k+1)} - C^{(k)}.$

 %%The vector $Sp^{(k)}$ indicates all possible rules in the system that could be applied at time $k.$  

\begin{remark} {\bf (Net gain vector)} {\rm \cite{zeng-etal2010}.} 
The product  $Sp^{(k)} \cdot M_{\Pi},$ denoted by $NG^{(k)},$ represents the net amount of spikes the system obtained at time $k$ from configuration $C^{(k)}.$ The sum $sp^{(k)} _1 b_{1j} + \cdots + sp^{(k)}_n b_{nj},$ $1\le k\le m,$ is the amount of spikes obtained in $\sigma_j$  at time $k.$  Thus $C^{(k+1)}(\sigma_j) = C^{(k)}(\sigma_j) + sp^{(k)} _1 b_{1j} + \cdots + sp^{(k)}_n b_{nj},$ $1\le k\le m,$ where $C^{(k)}(\sigma_j),$ and $C^{(k+1)}(\sigma_j)$ are the amount of spikes in $\sigma_j$ at time $k$ and $k+1,$ respectively.
 \end{remark}

Then we state the following important result from \cite{zeng-etal2010, adorna2019}:
 
%-------------------------------- Theorem Next configuration

\begin{theorem}\label{NextConfig} {\bf (Fundamental State Equation for SN P systems)} %{\rm \cite{zeng-etal2010}}
\\
Let $\Pi$ be an SN P system with total of  $n$ rules  and $m$ neurons. Let $C^{(k-1)}$ be given, then 
$$C^{(k)} = C^{(k-1)} + Sp^{(k-1)} \cdot M_{\Pi},$$
where  $Sp^{(k-1)}$ is the valid spiking vector with respect to $C^{(k-1)},$ and $M_{\Pi}$ is the spiking transition  matrix of $\Pi.$ 
\end{theorem}

\begin{definition}\label{ValidConfig} {\bf (Valid configuration)}\\
A configuration $C$ of some SN P system $\Pi$ is {\bf valid} if and only if $C=C^{(0)},$ or there exist valid configuration $C'$ and valid spiking vector $Sp',$ such that $C= C' + Sp' \cdot M_{\Pi},$  where $M_{\Pi}$ is the spiking transition matrix of SN P system $\Pi.$
\end{definition}

%Computation of SN P system $\Pi$ of Example \ref{example1} has been demonstrated in \cite{zeng-etal2010} to be faithfull as to how SN P system without delay works.

Using the formula from Theorem \ref{NextConfig}, we can obtain some valid configurations, $C^{(k)},$ from the initial configuration.

%------------------------------------------ Corollary C_k-by-C_0

\begin{corollary}\label{Ck-by-C0} {\rm \cite{adorna2019}}
Let $\Pi$ be an SN P system with total of  $n$ rules  and $m$ neurons.  Then at any time $k,$ we have
$$C^{(k)} = C^{(0)} + \left ( \displaystyle \sum_{i=0}^{k-1}Sp^{(i)} \right )\cdot M_{\Pi},$$ where $Sp^{(i)},$ $0\le i\le k-1,$ is a valid spiking vector.
\end{corollary}
 
The configuration of SN P system $\Pi$  is the state of $\Pi.$  This is the distribution of spikes among neurons in the system.   We use Corollary \ref{Ck-by-C0} to define reachability of a configuration $C$ of some SN P system $\Pi.$

\begin{definition} %{\bf (Reachability)}\\
Let $M_{\Pi}$ be the spiking transition  matrix of $\Pi,$ and $C^{(0)}$ its initial configuration.
A configuration $C$ is said to be {\bf $k$ reachable} in $\Pi$ if and only if there is in $\Pi$ a sequence of $k$ valid configurations $\{C^{(i)}\}_{i=0}^{k},$ for some $k,$ 
 such that $C^{(i)} = C^{(i-1)} + Sp^{(i-1)} \cdot M_{\Pi},$ 
 and $C^{(k)} =C.$   Moreover, $C$ must be valid configuration.
 
 We say the configuration $C$ is {\bf directly reachable} from $C^{(k)},$ for some positive integer $k$ if and only if there exists $Sp^{(k)},$ such that $C = C^{(k)} + Sp^{(k)} \cdot M_{\Pi}.$
 
 In particular, we define the set $$R_{\Pi}(C^{(0)}) = \{ C \mid C = C^{(0)} + \overline{s} \cdot M_{\Pi}, \text{where $\overline{s}$ is the sum of valid spiking vectors} \}$$ 
 the {\bf set of all reachable configuration $C$ from $C^{(0)}$} in $\Pi.$ 
 
\end{definition}

We note here that the vector $\overline{s}$ which is the sum of valid spiking vectors can be thought of as {\it sequence of valid spiking vectors} or {\it valid spiking sequence.}
%example 1
 
%-definition of reachable set $R_{\Pi}(C^{(0)}.$

%-definiton of direct reachability

%-------------------------------------------------------------------------------------
\section{Configuration graph $CG_{\Pi}$ of SN P system $\Pi$} \label{configgraph}
%------------------------------------------------------------------------------------- 
Let $\Pi$ be an SN P system without delay.   We describe a visual presentation of the computation of $\Pi$ as a directed graph.

\begin{definition}
Let $M_{\Pi}$ be the matrix representation of an SN P system $\Pi.$
A {\bf configuration graph} $CG_{\Pi}$ of $\Pi$ is the directed graph
$$CG_{\Pi} = (V,E),$$ where 
\begin{enumerate}
\item[] $V = \{C \mid C \in R_{\Pi} (C^{(0)}) \}$
\item[] $E = \{ (C^{(i)},C^{(j)}) \mid C^{(j)} = C^{(i)} + Sp^{(i)} \cdot M_{\Pi}, \text{ for some valid } Sp^{(i)} \}$
\end{enumerate}
\end{definition}

The configuration graph $CG_{\Pi}$ of an SN P system $\Pi$ visually represents the (valid) computations of $\Pi$ as directed paths from $C^{(0)}$ to some node or  configuration $C^{(k)},$ for some positive integer $k$ in $CG_{\Pi}.$   One can think of $CG_{\Pi}$  as a computation tree of some SN P system $\Pi.$

%------------------------------------------------
%EXample of Config graph
%-------------------------------------------------
\begin{example} We provide in Fig. 2 the configuration graph $CG_{\Pi}$ of the SN P system $\Pi$ of Example \ref{example1}.
\end{example}
\begin{figure}[h] \label{cg-snp1}
\begin{center}
\begin{tikzpicture}[node distance=2cm]

\node (c0) [neuron][] {(2,1,1) };
%\node (c1a) [neuron, below of=c0, label=left:$C^{(1)}$]{(1,0,2)};
\node (c1b) [neuron, below of=c0, xshift=2.5cm]{(2,1,2)};
%\node (c2a) [neuron, below of=c1a, label=left:$C^{(2)}$]{(1,0,0)};
\node (c2b) [neuron, below of=c1b, xshift=4cm, label=right:$C^{(1)}$]{$(2,1,2)$};
\node (c2c) [neuron, below of=c1b]{$(1,1,2)$};
\node (c3) [neuron, below of=c2c]{$(2,0,1)$};
\node (c4a) [neuron, below of=c3, xshift=-3.5cm]{$(0,1,1)$};
\node (c4b) [neuron, below of=c3, xshift=3.5cm]{$(1,1,1)$};
\node (c5) [neuron, below of=c4a]{$(1,0,1)$};
\node (c6) [neuron, below of=c5]{$(1,0,0)$};
\node (l1) [o, below of=c0, yshift=1cm, xshift=-1cm]{$(0,1,1,1,0)$};
\node (l2) [o, below of=c0, yshift=1.25cm, xshift=3.5cm]{$(1,0,1,1,0)$};
%\node (l3) [o, below of=c1a, yshift=1cm, xshift=-1cm]{$(0,0,0,0,1)$};
\node (l4) [o, below of=c1b, yshift=1cm, xshift=-1cm]{$(0,1,1,0,1)$};
\node (l5) [o, below of=c1b, yshift=1.25cm, xshift=3cm]{$(1,0,1,0,1)$};
\node (l6) [o, below of=c2c, yshift=1cm, xshift=-1cm]{$(0,0,1,0,1)$};
\node (l7) [o, below of=c3, yshift=1cm, xshift=-3cm]{$(0,1,0,1,0)$};
\node (l8) [o, below of=c3, yshift=1cm, xshift=.5cm]{$(1,0,0,1,0)$};
\node (l9) [o, below of=c3, yshift=1.45cm, xshift=4.5cm]{$(0,0,1,1,0)$};
\node (l10) [o, below of=c4a, yshift=1cm, xshift=-1cm]{$(0,0,1,1,0)$};
\node (l11) [o, below of=c5, yshift=1cm, xshift=-1cm]{$(0,0,0,1,0)$};

\node (la) [o, above of=c0, xshift=-2.5cm, yshift=-1cm]{$CG_{\Pi} \colon$};
\node (l1a) [o, left of=c0, xshift=-1.5cm]{$C^{(0)}$};
\node (l1b) [o, below of=c0, xshift=-3.5cm]{$C^{(1)}$};
\node (l1c) [o, below of=c0, yshift=-2cm, xshift=-3.5cm]{$C^{(2)}$};
\node (l1d) [o, below of=c0, yshift=-4cm, xshift=-3.5cm]{$C^{(3)}$};
\node (l1e) [o, below of=c0, yshift=-6cm, xshift=-3.5cm]{$C^{(4)}$};
\node (l1f) [o, below of=c0, yshift=-8cm, xshift=-3.5cm]{$C^{(5)}$};
\node (l1g) [o, below of=c0, yshift=-10cm, xshift=-3.5cm]{$C^{(6)}$};
%\node (sigma2) [neuron, right of=sigma3, xshift=2cm, label=right:$2$]{~~ $a$ ~~~ $a \rightarrow a$};

\draw [arrow] (c0) |- (c2c);
\draw [arrow] (c0) -- (c1b);
%\draw [arrow] (c1a) -- (c2a);
\draw [arrow] (c2b) |- (c1b);
\draw [arrow] (c1b) -- (c2b);
\draw [arrow] (c1b) -- (c2c);
\draw [arrow] (c2c) -- (c3);
\draw [arrow] (c3) -- (c4a);
\draw [arrow] (c3) -- (c4b);
\draw [arrow] (c4a) -- (c5);
\draw [arrow] (c5) -- (c6);
\draw [arrow] (c4b) |- (c3);

%\draw [arrow] (sigma2) -- (sigma3);
%\draw [arrow] (sigma3) -- (out);
\end{tikzpicture}
\end{center}
\caption{Configuration graph $CG_{\Pi}$ of the SN P system $\Pi$ that generates the set $\mathbb{N} -\{1\}.$}
\end{figure}
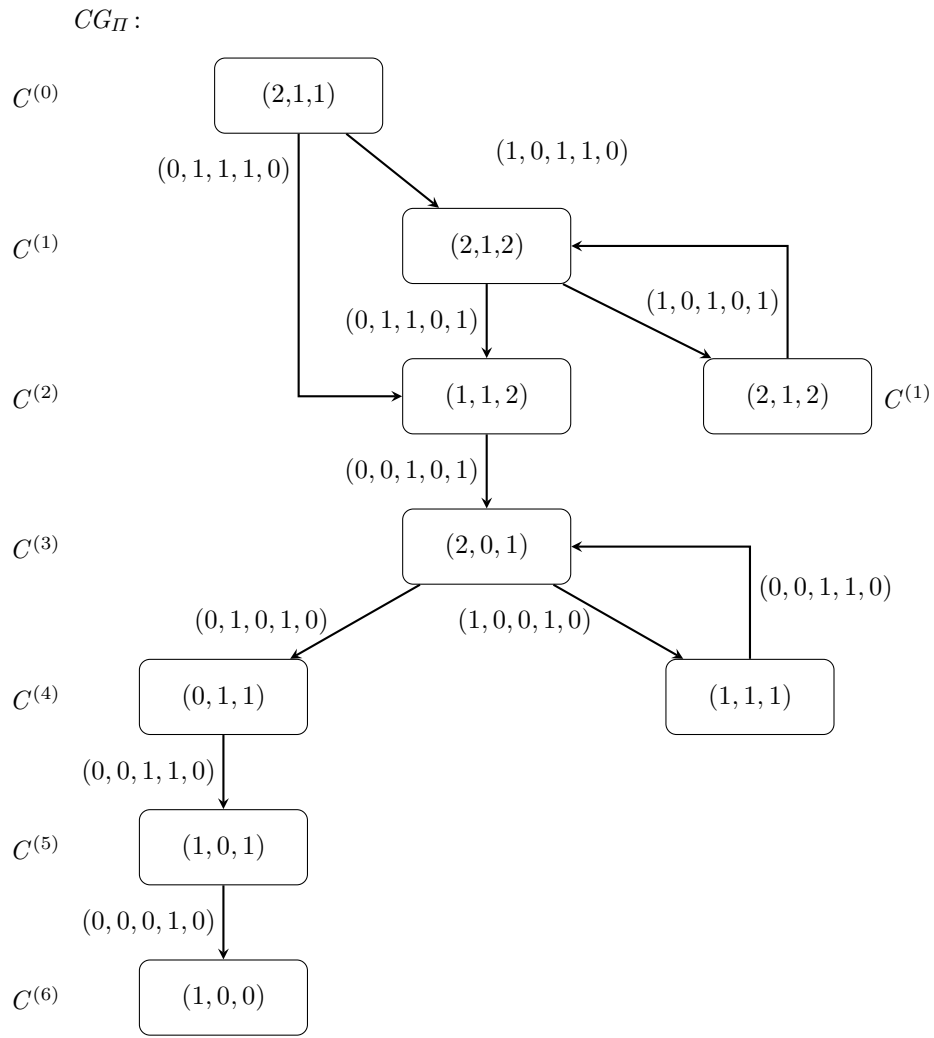

%\begin{definition}
%Let $\Pi$ be an SN P system.  Then the configuration graph $CG_{\Pi}$ of $\Pi$ is {\bf connected} if and only if for all configuration $C \in V(CG_{\Pi}),$ $C \in R_{\Pi}(C^{(0)}).$
%\end{definition}

\begin{observation} Given a  configuration graph $CG_{\Pi}$ of an SN P system $\Pi,$ we have the following:
\begin{enumerate}
\item $(C^{(i)}, C^{(j)}) \in E(CG_{\Pi})$ if and only if $C^{(j)}$ is {\bf  directly reachable} from $C^{(i)}.$
\item $CG_{\Pi}$ is a {\bf labelled (directed) graph,} where for every pair $C^{(i)}$ and  $C^{(j)}$ of vertices, $(C^{(i)}, C^{(j)})$ is labelled with the valid spiking vector $Sp^{(i)}.$  
\item The configuration graph $CG_{\Pi}$ of $\Pi$ is a {\bf one-component} or {\bf connected} directed graph 
\item $CG_{\Pi}$ is {\bf finite} if and only if $R_{\Pi}(C^{(0)})$ is finite.  Otherwise, $CG_{\Pi}$ is {\bf infinite.}
\end{enumerate} 
\end{observation}

\begin{definition}
Let $\Pi$ be an SN P system.  Then the configuration graph $CG_{\Pi}$ of $\Pi$ is {\bf strongly connected} if and only if for any two configurations $C, C' \in V(CG_{\Pi}),$ $C \in R_{\Pi}(C')$ and $C' \in R_{\Pi}(C).$
\end{definition}

\begin{observation}
Let $CG_{\Pi}$ be a strongly connected configuration graph of an SN P system  $\Pi.$ Then $CG_{\Pi}$ has a directed cycle..  
\end{observation}

%--------------------------------------------------------------------------
\section{Properties of SN P systems}\label{properties}
%--------------------------------------------------------------------------

We define below {\bf behavioral properties} of SN P systems.

\begin{definition}
We call a {\bf rule} $r$ {\bf live} for an initial configuration $C^{(0)}$ if for every $C^{(k)},$  $C^{(k)} \in R_{\Pi}(C^{(0)}),$  there exists a valid spiking sequence from $C^{(k)}$ that contains and applies $r.$ 

An SN P system $\Pi$ is {\bf live} for $C^{(0)}$ if all its rules are live for $C^{(0)}.$ 
\end{definition}

%Equivalently,

%\begin{definition}
%An SN P system $\Pi$ is {\bf live} if and only if for all $k,$ $k \geq 0, C^{(k)} \in R_{\Pi}(C^{(0)})$ and every rule $r$ in $\Pi,$ there exists a configuration $C^{(k')}$ such that $C^{(k')} = C^{(k)} + \overline{s} \cdot M_{\Pi},$ where $\overline{s}$ is a valid spiking sequence that contains and applies $r.$
%\end{definition}

This means that all rules in the SN P system $\Pi$ must not be useless permanently during the computation.
%%Thus an SN P system $\Pi$ being live would mean that it must not reach a point where at least one rule cannot be applied anymore in the computation.  
This is certainly not always the case of SN P systems especially, if an SN P system is generating or recognizing languages 
%%In these cases, 
where $\Pi$ must halt, which means no rules can be applied anymore.  
 This is equivalent to a so-called {\it deadlock} instance, that we can define as follows:
 %in the case of Petri nets.  Hence, 
 
 \begin{definition}
 A {\bf deadlock} in SN P system is a configuration where no rules can be applied.
 
 An SN P system is {\bf deadlock-free} for an initial configuration $C^{(0)},$ if no $C \in R_{\Pi}(C^{(0)})$ is a deadlock.
 \end{definition}
 
 We relax our definition of liveness for SN P systems and introduce the case of a so-called {\it non-dead} or {\it quasi-live} rule in SN P system.%transition in Petri nets.

\begin{definition}\label{def:quasi-live}
A rule $r$ is {\bf quasi-live} for an initial configuration $C^{(0)},$ if there is a valid spiking sequence from $C^{(0)}$ that contains and applies $r.$

We call an SN P system {\bf quasi-live} if all its rules are {\rm quasi-live}.
\end{definition}

Definition \ref{def:quasi-live}  of SN P systems liveness provides us allowance on some rules in the systems that may not be applied anymore after some time during the computation.

Most, if not all SN P systems so far constructed and reported in the literature are {\it not deadlock-free.} 
%%in particular, those SN P systems that halts after doing its job.   
Certainly, one could provide the system with a set of neurons that supply the system with spikes to avoid {\it ``deadlock''.}  Note that if $\Pi$ has a {\it ``feedback-loop''} mechanism %that could be represented as an iteration routing, 
then there is a possibility that $\Pi$ would be {\it ``live''.}

Next, we define a {\it bounded} SN P system with respect to the amount of spikes every neuron has after computation.  First, we define boundedness of a neuron.

\begin{definition}
A neuron $\sigma$ is {\bf bounded} for an initial configuration $C^{(0)}$ if there is a positive integer $s$ such that, $C(\sigma) \leq s,$ for every configuration $C \in R_{\Pi}( C^{(0)}),$ where $C(\sigma)$ is the amount of spikes in $\sigma$ in configuration $C.$  We say, $\sigma$ is an {\bf $s$-bounded neuron.}

An SN P system $\Pi$ is {\bf bounded} for an initial configuration $C^{(0)},$ if all neurons are bounded for $C^{(0)}.$   $\Pi$ is {\bf $s$-bounded} if all the neurons are $s$-bounded.

If $s=1,$ then we call the SN P system $\Pi,$ {\bf safe.}
\end{definition}

%Alternatively, we have 

%\begin{definition}
%An SN P systems $\Pi$ is {\bf bounded} if and only if for every neuron $\sigma$ in $\Pi,$ there exists a positive integer $s$ such that, $C(\sigma) \leq s,$ for every configuration $C \in R_{\Pi}(C^{(0)}),$ where $C(\sigma)$ is the amount of spikes in each $\sigma$ in configuration $C.$
%\end{definition}

%\begin{example}\label{bound}
%The SN P system of Example \ref{example1} is a $2$-bounded SN P system. 
%\end{example}

\begin{definition}
An SN P system $\Pi$ is {\bf reversible} if $C^{(0)} \in R_{\Pi}(C^{(k)}),$ for any $k \geq 0,$ where $C^{(k)} \in R_{\Pi}(C^{(0)}).$
\end{definition}

Now we define below properties of SN P system that do not depend on a fixed initial configuration.   These are called {\bf structural properties} of SN P systems.

\begin{definition}
An SN P system $\Pi$ in {\bf structurally live,} if there is an initial configuration $C^{(0)}$ such that $\Pi$ is live.
\end{definition}

\begin{definition}
An SN P system $\Pi$ is {\bf structurally bounded,} if it is bounded for any initial configuration $C^{(0)}.$ 
\end{definition}

\begin{definition}
An SN P system $\Pi$ is {\bf conservative,}  if there exists an $n$-vector $\overline{y}$ of positive integers such that for every initial configuration $C^{(0)}$ and for every configuration $C \in R_{\Pi}(C^{(0)}),$  we have $C \cdot \overline{y} = C^{(0)} \cdot \overline{y} =$ a constant.

If the $n$-vector $\overline{y}$ is allowed to be non-negative vector of integers, then $\Pi$ is called {\bf partial conservative.} 
\end{definition}

%---------------------------------------------------------
\section{SN P systems and its configuration graph}\label{snp+configgraph}
%---------------------------------------------------------
In this section, we shall consider properties of SN P systems that depends on the systems configurations (behavioral properties) and those that depends only on the connectivity of the neurons (structural properties).  We provide characterizations of these properties of SN P system $\Pi$ using it configuration graph $CG_{\Pi}.$ 

%-----------------------------------------------------
\subsection{Some behavioral properties}
%-----------------------------------------------------
\begin{theorem}Let $\Pi$ be an SN P systems without delay. $\Pi$ is {\rm bounded} if and only if  $CG_{\Pi}$  is finite. 
\end{theorem}

\proof  Let $\Pi$ be an SN P system without delay with $m$ neurons.   The proof will have two parts.

($\Longrightarrow$)  Let $\Pi$ be bounded, then there is a non-negative integer $s$ such that $C(\sigma) \leq s,$ for all neuron $\sigma$ in $\Pi.$  This means that each neuron $\sigma$  can have spikes equal to either $C(\sigma)=0, C(\sigma)=1, \ldots ,$ or  $C(\sigma)=s$ during computations.  Thus, we have only $(s+1)^m$ possible reachable configurations in $CG_{\Pi},$ since $\Pi$ has $m$ neurons. Therefore, $CG_{\Pi}$ is finite.  

($\Longleftarrow$) Let $CG_{\Pi}$ be finite.  This means that $V(CG_{\Pi})$ is finite if and only if $R_{\Pi}(C^{(0)})$ is finite.  Let $R_{\Pi}(C^{(0)})$ has $k$ elements.  Then any configuration can be reached by at most $k-1$ valid spiking sequences from $C^{(0)}.$  Since the amount of spikes in each neuron increases by at most $p$ spikes for each spiking sequence, where $p$ is the maximum spikes produced by any rule in $\Pi.$  Then for every $\sigma$ in $\Pi,$ $C(\sigma)$ is bounded above by $C^{(0)}(\sigma) + p(k-1).$   Therefore, $\Pi$ is bounded.

\myqed

\begin{theorem}Let $\Pi$ be an SN P systems without delay. $\Pi$ is {\rm deadlock-free} if and only if  all of the vertices in $CG_{\Pi}$  have at least one synapse going out. \end{theorem}

\proof Let $\Pi$ be an SN P systems without delay. We proceed as follows:

($\Longrightarrow$)  Suppose there is a $C \in V(CG_{\Pi}).$ Let us call the number of synapse going out of a node or configuration$C$ in $CG_{\Pi}$ an $outdeg(C).$  Let $outdeg(C) =0.$ This means that after reaching $C$ from $C^{(0)}$ the amount of spikes in each neuron at configuration $C$ is not enough to fire any of the rules in every neuron, and therefore reached a deadlock.  This a contradiction, since $\Pi$ is deadlock-free.  Therefore all vertices or configurations $C$ in $CG_{\Pi}$ has $outdeg(C) \geq 1.$

($\Longleftarrow$) Since each configuration $C$ in $CG_{\Pi}$ has $outdeg(C) \geq 1,$ then clearly that every configuration reached allows to have a valid spiking vector to transition from one configuration to the next and so on. Hence, $\Pi$ never reached a configuration where its computation stops or no more rule to apply.   Therefore, $\Pi$ is deadlock-free.

\myqed
 
\begin{theorem}Let $\Pi$ be an SN P systems without delay. $\Pi$ is {\rm live} if and only if  for each vertex $C$ of $CG_{\Pi},$ there exists a path from $C$ to $C^{(k)},$ for some positive integer $k,$ such that the labels of the edges of the path indicate that they contain and apply all the rules of $\Pi.$  \end{theorem}

\proof Let $\Pi$ be a live SN P system without delay.  

($\Longrightarrow$) Let $CG_{\Pi}$ be the configuration graph of $\Pi.$  Note that vertices of $CG_{\Pi}$ are all reachable configurations of $\Pi.$  Let $C=C^{(i)},$ for some integer $i$ such that $C^{(i)} \in R_{\Pi}(C^{(0)}).$   Let $r_1, r_2, r_3, \ldots , r_n$ be all the rules in $\Pi.$    Since $\Pi$ is live, we can find a configuration from $C^{(i)}$ that enables at least a rule among the $n$ rules of $\Pi.$  Let $r_1$ be applicable with respect to $C^{(i)}.$ Then $C^{(i+1)} = C^{(i)} + Sp^{(i)} \cdot M_{\Pi},$ for some valid spiking vector $Sp^{(i)}$ that enables $r_1.$ With respect to $C^{(i+1)},$ let $Sp^{(i+1)}$ be valid spiking vector that enables, say, at least $r_2.$ Then we have $C^{(i+2)} = C^{(i+1)} + Sp^{(i+1)} \cdot M_{\Pi},$ and so on.  This repeated applications of valid spiking vectors obtained from every succeeding configuration can be done due to liveness property of $\Pi.$ This application will eventualy lead us to a sequence of valid spiking vectors that enables all the rules of $\Pi.$  Then finally reach some configuration $C^{(k)},$ for some positive integer $k,$ such that $C^{(k)} = C^{(k-1)} + Sp^{(k-1)} \cdot M_{\Pi}$  where $C^{(k)} \in R_{\Pi}(C^{(i)}).$  It can be realized that $C^{(k)} = C^{(i)} + (Sp^{(i)} + Sp^{(i+1)} + \cdots + Sp^{(k-1)}) \cdot M_{\Pi}.$ And from $CG_{\Pi},$ we can identify a labelled directed path from $C^{(i)}$ to $C^{(k)},$ labelled sequentially by the following valid spiking vectors $Sp^{(i)}, Sp^{(i+1)} , \ldots, \text{ and } Sp^{(k-1)},$ which enable  and apply all the $n$ rules in $\Pi.$  

($\Longleftarrow$) All vertices of $CG_{\Pi}$ represents the reachable configurations of $\Pi$ from $C^{(0)}.$  Suppose $C^{(i)}$ is a vertex of $CG_{\Pi},$ for some positive integer $i,$ such that we can follow a labelled directed path from $C^{(i)}$ to a vertex $C^{(k)}$ for some positive integer $k.$  The labels are valid spiking vectors that indicate applicable rules of $\Pi,$ for every succeeding  configuration. This implies $C^{(k)}= C^{(i)} + \overline{x} \cdot M_{\Pi},$ where $\overline{x}$ is the sum of sequence of valid spiking vectors obtained starting from $C^{(i)}$ up to $C^{(k-1)}.$  Moreover, all entries in $\overline{x}$ are non-zero.   Therefore, all rule of $\Pi$ are applied in $\overline{x},$ that implies  liveness of $\Pi.$       

\myqed

%\begin{theorem}Let $\Pi$ be an SN P systems without delay. $\Pi$ is {\rm periodic with respect to configurations,} that is,  if and only if $CG_{\Pi}$ represents a strongly connceted $\Pi.$\end{theorem}

%\proof The proof will have two parts.

%($\Longrightarrow$) 

%($\Longleftarrow$)

%\myqed

\begin{theorem}
Let $\Pi$ be an SN P system without delay.  Then 
$\Pi$ is {\rm reversible} if and only if  $CG_{\Pi}$ is strongly connected.
\end{theorem}

\proof Let $\Pi$ be an SN P system without delay.

($\Longrightarrow$)  Suppose $\Pi$ is reversible.   Then for any positive integer $k,$ all reachable configuration $C^{(k)}$ from $C^{(0)},$ have directed path leading to $C^{(0)}$ from $C^{(k)}.$   Suppose for any positive integer $k',$ $C^{(k')} \in R_{\Pi}(C^{(0)}),$ such that $k' \ne k.$ Since $\Pi$ is reversible, $C^{(0)} \in R_{\Pi}(C^{(k')}).$ Thus we can find a directed path in $CG_{\Pi}$ from $C^{(k')}$ to $C^{(k)}.$  One can take the path drawn by the fact that $C^{(0)} \in R_{\Pi}(C^{(k)}),$ followed by the path from $C^{(0)}$ to $C^{(k')}.$  Hence, for any positive integers $k$ and $k',$ we can find directed path from $C^{(k)}$ to $C^{(k')}$ and back.
Therefore, $CG_{\Pi}$ is strongly connected.

($\Longleftarrow$)  Let $CG_{\Pi}$ be strongly connected.  Then  $C^{(0)} \in R_{\Pi}(C^{(k)}),$ for any positive integer $k.$ Therefore, $\Pi$ is reversible.

\myqed

%----------------------------------------------------
\subsection{Some structural properties}
%----------------------------------------------------
We now look at some structural properties of SN P system $\Pi.$  These are properties of $\Pi$ that are not dependent on $C^{(0)}.$  

\begin{theorem}
Every live SN P system $\Pi$ is structurally live.
\end{theorem}
\proof  This follows from the definition.
\myqed

To prove Theorem \ref{structural-bdd} below (by contradiction), we will use the result called Minkowski-Farkas Lemma \cite{farkas1902, gale1951}: Let $A$ be an $n \times m$ matrix over $\mathbb{R},$ $\overline{x}$ is a real column vector of size $n$ and $\overline{b}$ is a real column vector of size $m.$  The linear system $A \overline{x} = \overline{b},  \overline{x} \geq 0$ has a solution if and only if for all $\overline{y} \in \mathbb{R}^m,$ $A^T \cdot \overline{y} \geq 0,$ we have $\overline{y} \geq 0$ and $\overline{b} \geq 0.$ 

\begin{theorem}\label{structural-bdd}
An SN P system $\Pi$ without delay is structurally bounded if and only if there exists a positive column vector $\overline{y}$  of integers of size $n$ such that $M_{\Pi} \cdot \overline{y} \leq 0.$ 
\end{theorem}

\proof  Let $\Pi$ be an SN P system without delay.

($\Longrightarrow$)
Suppose we have a positive column vector $\overline{y}$ of integers of size $n,$ such that $M_{\Pi} \cdot \overline{y} > 0.$  Then by Minkowski-Farkas Lemma, there exists a row vector $\overline{x} \geq 0$ of integers of size $m,$ such that $\overline{x} \cdot M_{\Pi} >0.$
Then we can find configurations $C$ and $C^{(0)},$ such that $C-C^{(0)} = \overline{x} \cdot M_{\Pi} > 0$ or $C > C^{(0)}.$  We choose $C^{(0)}$ large enough to ensure a sequence of valid spiking vectors, $\overline{s},$  such that $\overline{s}=\overline{x}$ can be repeated indefinitely.  Thus $\Pi$ is unbounded.

($\Longleftarrow$)  Suppose there exists a positive column vector $\overline{y}$ of integers of size $n,$ such that $M_{\Pi} \cdot \overline{y} \leq 0.$  Let $C \in R_{\Pi}(C^{(0)}),$ where $C^{(0)}$ is some initial configuration of $\Pi.$  This implies that $C = C^{(0)} + \overline{s} \cdot M_{\Pi},$ where $\overline{s} \geq 0.$   Then we obtain the following sum of products $$C \cdot \overline{y} = C^{(0)} \cdot \overline{y} + \overline{s} \cdot M_{\Pi} \cdot \overline{y}.$$ Since $M_{\Pi} \cdot \overline{y} \leq 0$ and $\overline{s} \geq 0,$ $$C \cdot \overline{y} \leq  C^{(0)} \cdot \overline{y}.$$  Let $C(\sigma_i)$ be the number of spikes in $\sigma_i$ at configuration $C$ for some neuron  $\sigma_i$ in $\Pi.$    Let $\overline{y} (\sigma_i)$ be the $i^{th}$ entry of $\overline{y}.$  Then $$C(\sigma_i) \leq \frac{C^{(0)} \cdot \overline{y}}{\overline{y}(\sigma_i)}.$$
Thus, $C(\sigma_i)$ is bounded for each $\sigma.$ 
\myqed

\begin{theorem}
An SN P system $\Pi$ without delay is {\bf conservative} if and only if there exists a positive column vector $\overline{y}$ of integers of size $n,$ such that $$M_{\Pi} \cdot \overline{y} = {\bf 0}.$$
\end{theorem}
\proof Let $\Pi$ SN P system without delay.

($\Longrightarrow$)   Let $\Pi$ be conservative.  Then by definition, there exists a positive column vector $\overline{y}$ of integers of size $n,$ such that for every initial configuration $C^{(0)}$ and for every configuration $C \in R_{\Pi}(C^{(0)},$  we have $C \cdot \overline{y} = C^{0)} \cdot \overline{y}$ is a constant.  Notice that $C \cdot \overline{y} = C^{(0)} \cdot \overline{y} + \overline{s} \cdot M_{\Pi} \cdot \overline{y},$ for any positive column vector $\overline{y}$ of integers of size $n.$   Since $\Pi$ is conservative, therefore,  $$M_{\Pi} \cdot \overline{y}={\bf 0},$$ since $\overline{s} \geq 0.$

($\Longleftarrow$) Suppose we have a positive column vector $\overline{y}$ of integers of size $n,$ such that $M_{\Pi} \cdot \overline{y} = {\bf 0}.$  Then for every $C^{(0)}$ and every configuration $C \in R_{\Pi}(C^{(0)},$ we have, $$C \cdot \overline{y} = C^{(0)} \cdot \overline{y},$$  as required.   Therefore, $\Pi$ is conservative.
\myqed

\begin{corollary}
An SN P system $\Pi$ without delay is {\bf partial conservative} if and only if there exists a positive column vector $\overline{y}$ of integers of size $n,$ such that $$M_{\Pi} \cdot \overline{y} = {\bf 0}.$$
\end{corollary}

%-----------------------------------
\section{Final Remarks}\label{finalremarks}
%-----------------------------------

We have demonstrated that the properties of SN P system without delay mimic the behavioral and structural properties of a (place/transition) Petri net  \cite{david-etal2010, reisig1998, murata1989}.  We believe that several other properties of Petri nets could be interpreted as properties of SN P systems (see \cite{marian2020}). It is not hard to see that the converse of the definitions of the structural  properties of SN P system as defined in Section \ref{properties} may not always be true.

The matrix representation defined in \cite{zeng-etal2010, adorna2019} could be used to find other possible algebraic properties of SN  P systems.  Note when computing backwards \cite{naranjo2010} the fundamental state equation (\ref{eq1}) is found useful. Solutions to the fundamental state equation for SN P system connote reachability of configurations.  

Since SN P systems and their configuration graphs are directed graphs or digraphs, we could investigate further properties of these digraphs and perhaps we can observe, if any, their implications to the behavioral and structural properties of SN P systems.  The techniques from \cite{frost1992, norman1965} on digraphs could be useful in possibly proving some properties of SN P systems using matrices.   A good reference on the results on digraphs could be \cite{bang2009}.

Also, we have demonstrated that strong connectedness is related to reversibility of $\Pi.$  This may be relevant to periodicity and other possible behavioral and structural invariance in $\Pi.$

Finally, we suggest to explore further the behavioral and structural properties not only of SN P systems and variants but also the other types of P systems and their variants.   We could either use the computation tree s in \cite{naranjo2010} or adopt our configuration graph for various types of P systems.

%In this preliminary investigation, we have demonstrated properties of SN P systems mimic behavioral and structural properties of Petri nets.   Note that Petri nets are defined structurally as directed bipartite graphs while SNP systems are directed graphs not necessarilly bipartite. In this paper we mapped to marked Petri net an SN P system with {\it (non-empty) initial configuration,} while an SN P system with {\it $\emptyset$-configuration,} that is there is no initial spike found in all neurons,  cis mapped to an unmarked Petri net. 

%Also, several other properties of Petri nets \cite{david-etal2010, reisig1998, murata1989} could be interpreted as properties of SN P systems. 

%------------------------------------------------
\subsection*{Acknowledgement}
%------------------------------------------------
The author would like to thank the support from DOST-ERDT research grants; Semirara Mining Corp. Professorial Chair for Computer Science of College of Engineering, UPDiliman; RLC grant from UPD-OVCRD.

%---------------------------------------------

%\bibliographystyle{splncs03}
%\bibliography{references}

\end{document}